\documentclass[10pt, conference]{IEEEtran}

\usepackage{cite}
\usepackage{amsmath,amssymb,amsfonts}
\usepackage{algorithmic}
\usepackage{graphicx}
\usepackage{array}
\graphicspath{{./figures/}}
\usepackage{textcomp}

\usepackage[nolist]{acronym}
\begin{acronym}[DEN-ng]
\acro{DNN}{Deep Neural Network}
\acro{DL}{Deep Learning}
\acro{GPU}{Graphics Processing Unit}
\acro{CPU}{Central Processing Unit}
\acro{BSQ}{Bounding Sphere Quantization}
\acro{EM}{Expectation-Maximization}
\acro{SGD}{Stochastic Gradient Descent}

\end{acronym}

\usepackage{multirow}
\usepackage{color, colortbl}

\usepackage{lipsum}

\usepackage[caption=false,font=footnotesize]{subfig}

\def\BibTeX{{\rm B\kern-.05em{\sc i\kern-.025em b}\kern-.08em
		T\kern-.1667em\lower.7ex\hbox{E}\kern-.125emX}}

\newcommand{\kmeans}{\textit{k}-Means}

\newcolumntype{R}[1]{>{\raggedleft\let\newline\\\arraybackslash\hspace{0pt}}r{#1}}

\begin{document}
	
	\title{Neural Network-based Quantization\\for Network Automation}

	\author{
		\IEEEauthorblockN{
			M{\'a}rton Kaj{\'o}\IEEEauthorrefmark{1},
			Stephen S. Mwanje\IEEEauthorrefmark{2},
			Benedek Schultz\IEEEauthorrefmark{2},
			Georg Carle\IEEEauthorrefmark{1}
		}
		\IEEEauthorblockA{
			\IEEEauthorrefmark{1}Technical University of Munich, Department of Informatics\\
			Email: \{kajo, carle\}@net.in.tum.de
		}
		\IEEEauthorblockA{
			\IEEEauthorrefmark{2}Nokia Bell Labs\\
			Email: stephen.mwanje@nokia-bell-labs.com, \\
			       schultz.benedek@gmail.com
		}
	}
	
	\maketitle
	
	\begin{abstract}
		Deep Learning methods have been adopted in mobile networks, especially for network management automation where they provide means for advanced machine cognition. Deep learning methods utilize cutting-edge hardware and software tools, allowing complex cognitive algorithms to be developed. In a recent paper, we introduced the \ac{BSQ} algorithm, a modification of the \kmeans{} algorithm, that was shown to create better quantizations for certain network management use-cases, such as anomaly detection. However, \ac{BSQ} required a significantly longer time to train than \kmeans{}, a challenge which can be overcome with a neural network-based implementation. In this paper, we present such an implementation of \ac{BSQ} that utilizes state-of-the-art deep learning tools to achieve a competitive training speed.	
	\end{abstract}
	
	
	\acresetall
	
	\section{Introduction}
	
		Recently, many technological fields have adopted \ac{DL}, or more specifically: \acp{DNN}. \acp{DNN} are notorious both for their capacity to solve complex problems, and for the extraordinary computational power required to train them. To keep training times sensible (and make inference fast), \ac{DL}-specific software frameworks were developed, such as PyTorch\footnote{https://www.pytorch.org/} or TensorFlow\footnote{https://www.tensorflow.org/}, which are capable of utilizing dedicated hardware accelerators, \acp{GPU}. As deep learning is envisioned to be a major part of everyday life, future mobile networks are also planned to include these dedicated hardware accelerators \cite{DLwirelessSurvey}, to allow for deep-learning enabled applications and management functions.
		
		Network management automation is no exception to the deep learning trend, with an aim towards the realization of highly intelligent, cognitive management tasks \cite{MwanjeCAN}. As there is planned hardware support for deep learning, it makes sense to also utilize this power for the management of the future network. For now, deep learning is heavily investigated, and there is a rapid adoption of the deep learning infrastructure among network automation researchers, both hardware and software \cite{zappone2019wireless}.		

		Deep learning is mostly successful in supervised learning tasks - those involving a clearly defined, finite problem, where training data with the expected output (labels) attached is readily available for all foreseeable circumstances. Network management, however, often involves semi- or unsupervised learning scenarios - where the expected output is not available in the training data, and machine learning is specifically needed to bring intuition/agency into the automation, to make the system capable of reacting to unforeseen circumstances.

		One such often used task is quantization (clustering). Quantization, similar to other unsupervised learning tasks, has not seen much advancement as the supervised learning tasks. Older methods are still used today, implemented with aging software packages that can not easily integrate into \ac{DL} frameworks or leverage available hardware acceleration. This paper serves a follow-up to \cite{kajo2018equal}, where a modification of the well-known \kmeans{} quantization algorithm called \ac{BSQ} was introduced. \ac{BSQ} was shown to behave better than \kmeans{} in particular network-management oriented tasks, such as anomaly detection, or network-state mapping. However, one major disadvantage compared to \kmeans{} was the significantly increased processing time, which was, at the time, left for future study. In this paper, we show a way of implementing both \kmeans{} and \ac{BSQ} in a way that is well-aligned with modern \ac{DL} frameworks, and evaluate the performance against a readily available common \kmeans{} implementation.
	
	\section{\kmeans{} and \ac{BSQ}}

		We consider two quantization algorithms in this paper; the traditional Lloyd's \kmeans{} algorithm \cite{lloyd1982least}, and our modification thereof called \ac{BSQ}. In their original form, both algorithms use the \ac{EM} iterative logic for fitting the quanta. In this paper, both algorithms are reformulated into a neural network-style logic, with operations organized into layers, and \ac{SGD} being used for fitting instead of \ac{EM}. The two algorithms will be discussed together, as they share many aspects and implementation details.		
		
		Quantization divides the input space into finite volumes. \kmeans{} and BSQ does this by iteratively moving $k$ quantum centerpoint (centroid) locations so that a certain measure of goodness of fit, calculated on the whole training dataset, is minimized. This can be viewed as the \textit{global} (training-set-wide) optimization target. For \kmeans{}, this goodness of fit is calculated as the average of distances between training points and their closest centroids. To find a minimum, the original algorithm alternates two steps: 
		\begin{itemize}
			\item \textit{Expectation}: The training points are assigned to the closest quantum centroid.
			\item \textit{Maximization}: The quantum centroids are moved to best fit the assigned training points.			
		\end{itemize}
	
		By moving the quantum centroids to fit the assigned points, \kmeans{} fulfills a \textit{local} (quantum-wide) optimization goal. The algorithm does not have an explicit mechanism to reach the global optimization target, however, the interactions between the assignment and the localized optimizations moves the whole quantization towards the global goal. The distance measure used is usually the $L_2$ (Euclidean distance, or $2$-norm from the group of $p$-norms), in which case the local optimum for each centroid is the mean of the assigned points, hence the name \kmeans{}.
		
\begin{figure}[h]
			\vspace{-0.2in}
			\centering
			\subfloat[\kmeans{}]{
				\includegraphics[width=0.46\linewidth]{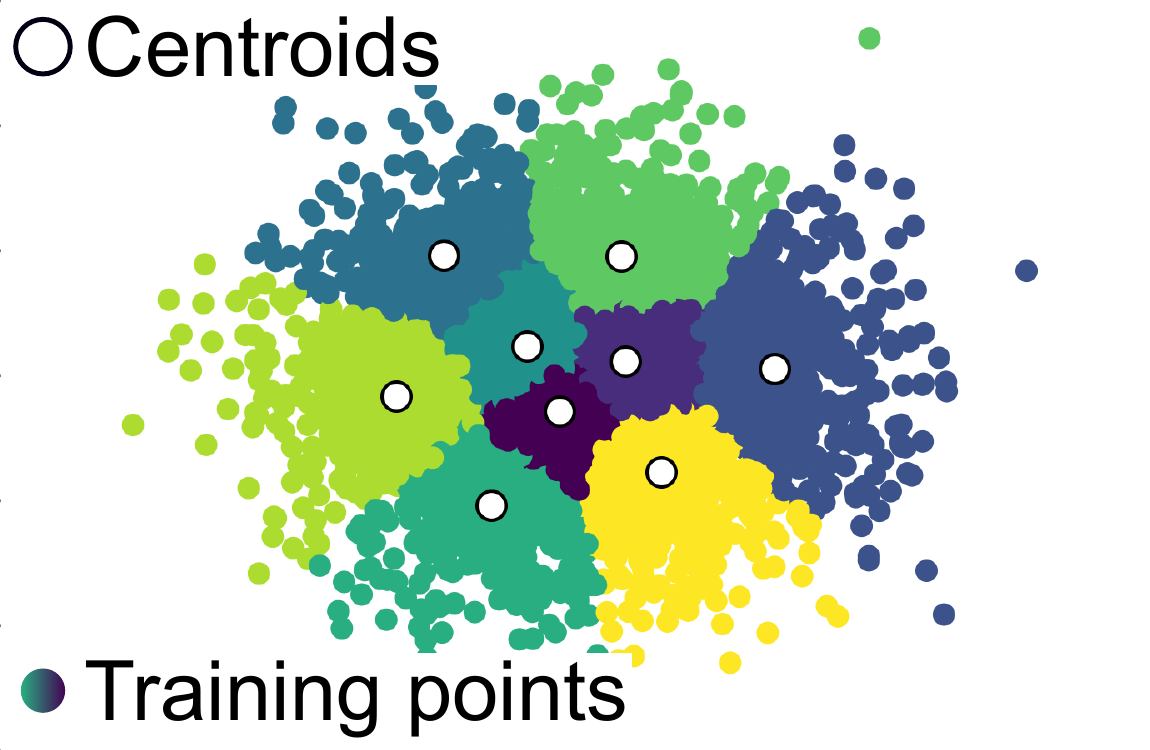}
			}
			\subfloat[\ac{BSQ}]{
				\includegraphics[width=0.46\linewidth]{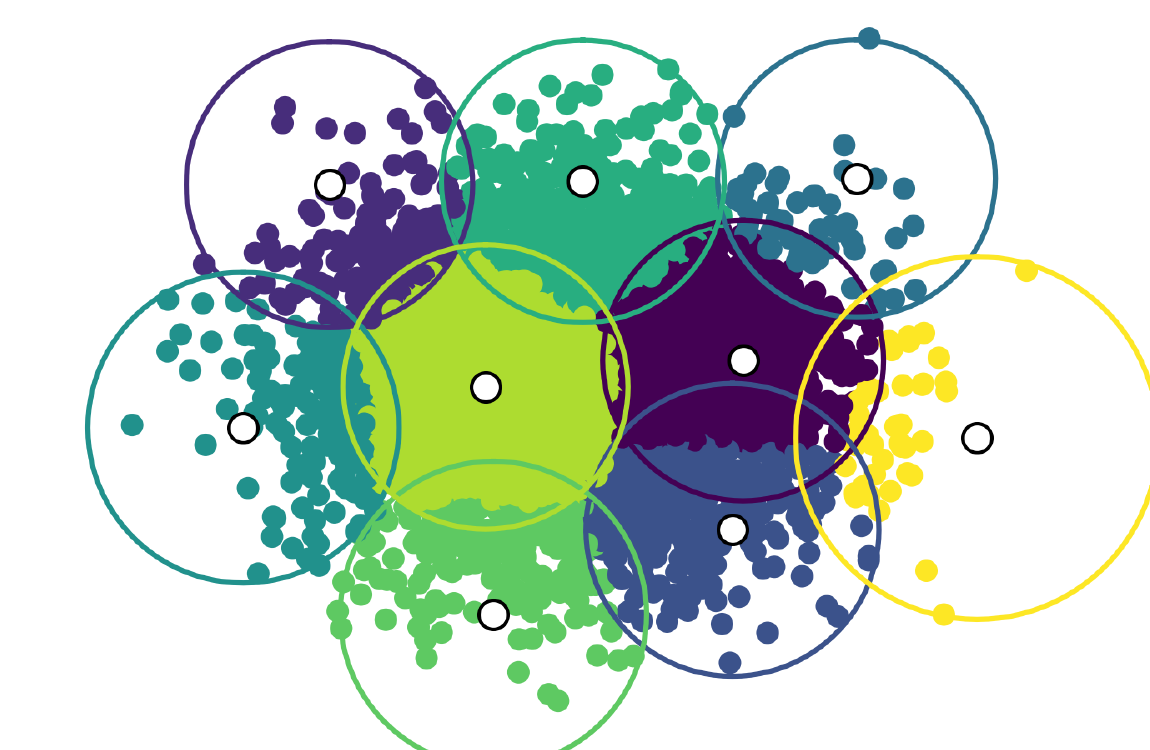}
			}
			\caption{Illustration of the difference between \kmeans{} and \ac{BSQ}}
			\label{fig:illustration}
			\vspace{-0.1in}
		\end{figure}

		\ac{BSQ} changes the local optimization target. Instead of trying to minimize the average of all distances within a quantum, \ac{BSQ} tries to minimize only the single largest distance. To achieve this, each centroid is moved to the center of the \textit{minimal bounding sphere} \cite{bounding} of the assigned points in the maximization steps. This change in the local optimization target also changes the global behavior; \ac{BSQ} fits quantizations where the maximum distance in the whole quantization is minimal. This makes \ac{BSQ} tend towards quanta with equal assigned volumes, which has been shown in \cite{kajo2018equal} to be preferable over \kmeans{} in tasks such as anomaly detection, where the goal is to chart/explore the input space, without necessarily following the density distribution of the training points. Visually, BSQ distributes quanta evenly in the training space, producing quanta with equal sizes, whereas \kmeans{} focuses more on densely populated parts, covering sparsely populated parts with larger quanta. An illustration of this difference in behavior can be seen on Fig. \ref{fig:illustration}.
		
	\section{Neural network implementation}
		
\begin{figure}[h]
			\centering
			\includegraphics[width=\linewidth]{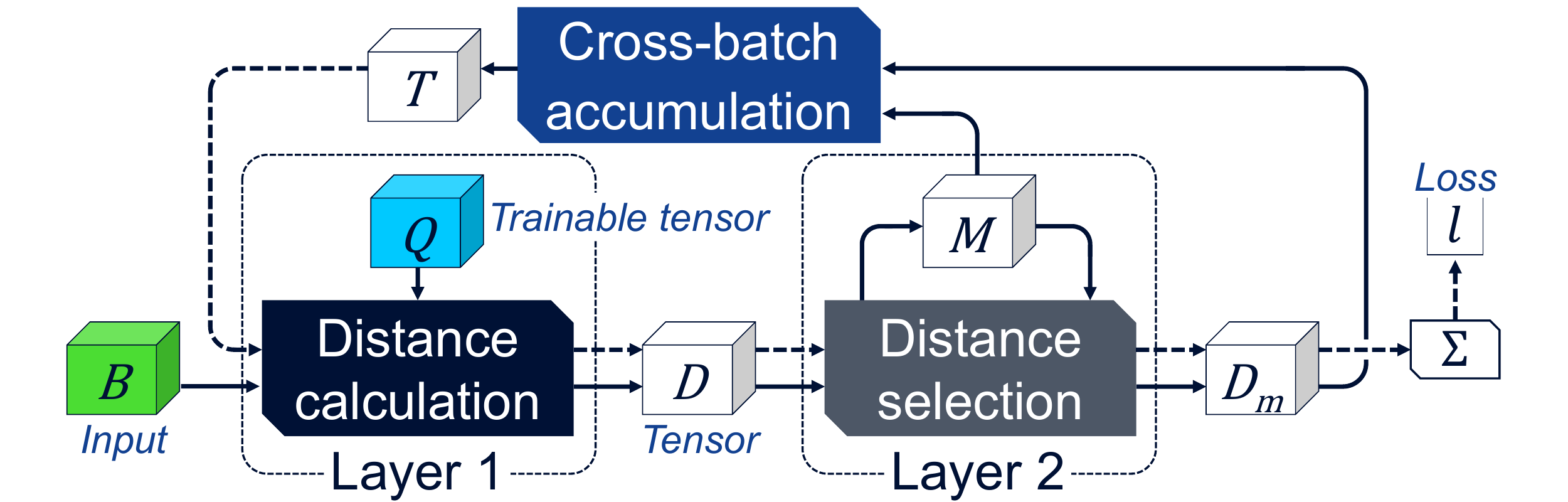}
			\caption{Overview of the processing steps for \kmeans{} and \ac{BSQ} implemented as a neural-network}
			\label{fig:quant_overview}
			\vspace{-0.15in}
		\end{figure}

		The biggest change in moving the \kmeans{} and \ac{BSQ} algorithms to a neural network-style logic is the switch from the \ac{EM} optimization to Stochastic (randomly permuted and batched) Gradient Descent. For \ac{SGD} to work, the distance calculations need to produce a single loss value, that is to be backpropagated to update the quantum centroids. Selecting which of the distances between quantum centroids and training points contribute to the loss value differentiates between \kmeans{} and \ac{BSQ}. Both the distance calculation and the distance selection can be realized as neural network layers. Additionally, the stochastic nature of batching breaks \ac{BSQ}, so a cross-batch accumulation is required. This accumulation, however, also benefits \kmeans{}. An overview the whole process can be seen on Fig. \ref{fig:quant_overview}, whose individual steps are discussed in the following sections.
		
		\subsection{Distance calculation layer}

			The core of \kmeans{}-like quantization is a calculation of distance, measured between training points and quantum centroids. For this paper, we consider $p$-norms only, as these cover the most commonly used distances. PyTorch and TensorFlow includes ready implementations of calculating $p$-norms of vectors organized into tensors (multi-dimensional matrices), but complete functions to calculate set-to-set distances between two set of points are missing from both libraries. To overcome this, \textit{broadcasting}, a technique that is available in both libraries can be used, which enables the calculation of all set-to-set distances without the need to manually duplicate data in memory.
			
\begin{figure}[h]
				\vspace{-0.1in}
				\centering
				\includegraphics[width=\linewidth]{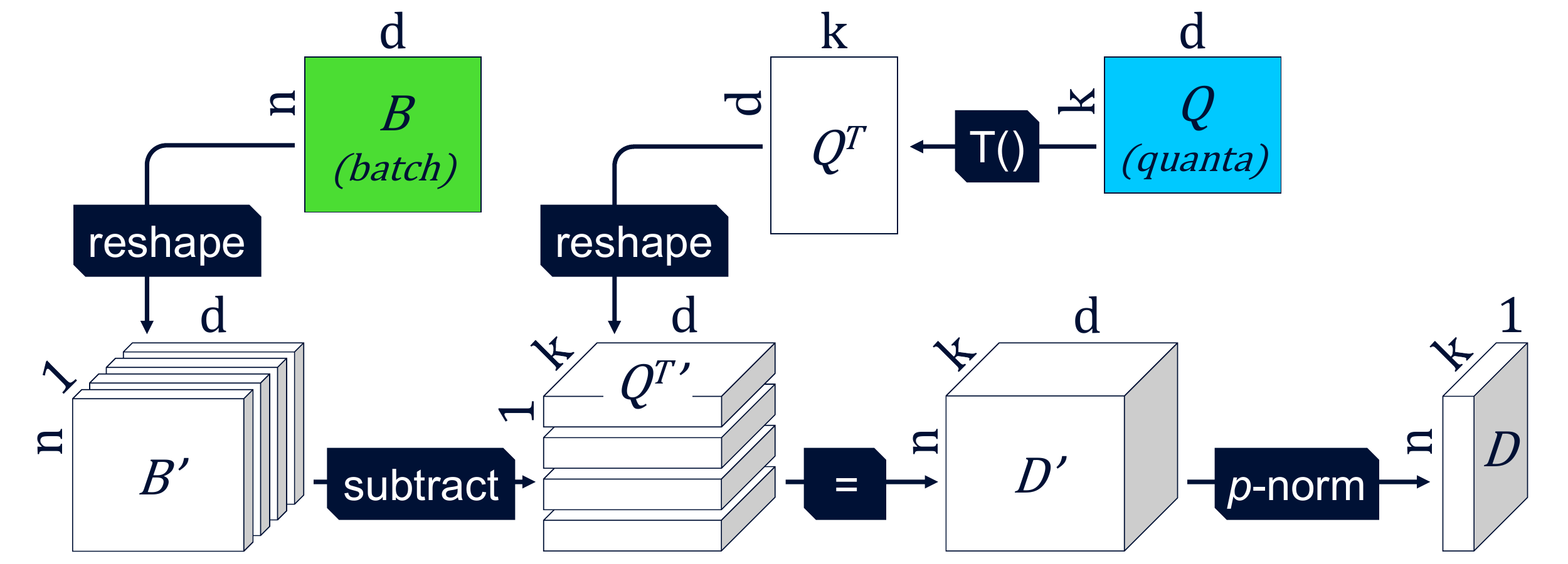}
				\caption{Distance calculation steps}
				\label{fig:distance_calc}
				\vspace{-0.15in}
			\end{figure}

			Let $B$ (batch) be a tensor of shape $(n\: rows, d\: columns)$ containing training points, where $n$ is the size of the current batch, and $d$ is the number of dimensions. Let $Q$ be a tensor of shape $(k, d)$ containing $k$ quantum centroids. In this case, $B$ can be recast to shape $(n, d, 1)$ resulting in tensor $B'$, and $Q{^T}$ (the transpose of $Q$) can be recast to shape $(1, d, k)$ resulting in tensor $Q{^T}'$ without any memory copies created. The tensor dimensions of size $1$ can then be reused without copy in the element-wise subtraction $B' - Q{^T}' = D'$. The resulting tensor $D'$ with shape $(n, d, k)$ contains all pairwise difference vectors between $B$ and $Q$. Finally, the pairwise distances between $B$ and $Q$ can be calculated by computing the $p$-norm of $D'$ in the direction of the middle tensor dimension of size $(d)$, reducing $D'$ into $D$ with shape $(n, k)$. The whole process can be seen on Fig. \ref{fig:distance_calc}.
			
		\subsection{Distance selection layer}
		
\begin{figure}[h]
				\vspace{-0.1in}
				\centering
				\includegraphics[width=\linewidth]{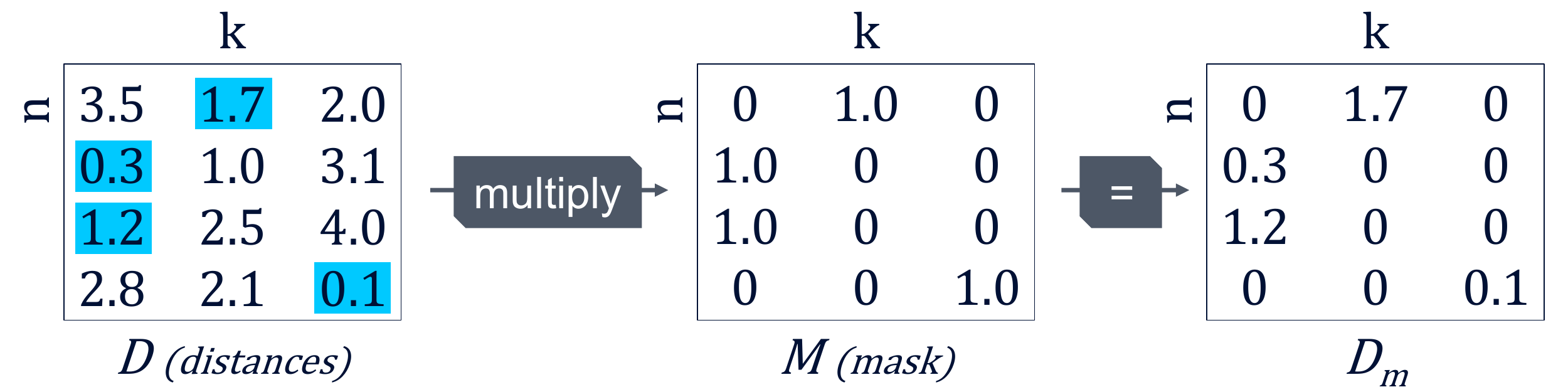}
				\caption{Distance selection through masking}
				\label{fig:distance_select}
				\vspace{-0.15in}
			\end{figure}

			For both algorithms, only the distances to the closest quantum centroid should contribute to the final loss value. This translates into the need of selecting the smallest distance for each training point, which is a row-wise minimum search in tensor $D$. However, cross-batch accumulation needs to retain information about which distance belongs to which quantum, so instead of selecting the smallest values, it is better to mask all other unimportant distance values by multiplying them with $0$. To do this, the masking tensor $M$ of shape $(n, k)$ is created, which contains $1$-s at places where $D$ contains row-wise minima, and $0$-s everywhere else. Element-wise multiplying $D * M = D_m$ results in the masked distance tensor $D_m$ with shape $(n, k)$. The operation can be seen on Fig. \ref{fig:distance_select}.

		\subsection{Cross-batch accumulation}
		
			To be clear, the step of cross-batch accumulation is not necessary for \kmeans{}. Because a large-enough random sample from a set of points retains the distribution of the original set with a high confidence, the stochastic samples contained within the batches likely have the same mean as the whole set of training points (weak law of large numbers \cite{lawOfLarge}). Because of this, for \kmeans{} it is enough to calculate the mean of the $D_m$ tensor and backpropagate this value as the final loss in every iteration.
			
			However, this is not true for \ac{BSQ}. Finding the farthest point in each batch for each quanta, and trying to minimize those distances will not result in a similar behavior as finding the farthest points in the whole training set. To overcome this, we accumulate the optimization targets across batches, and only update quantum centroids after a certain number of batches were processed. The number of batches to be processed before each update is the user-set parameter $r$. In case of $r = n_{batches}$, there is no accumulation (updates happen at every batch), whereas for $r = 1$, updates only happen after all batches were accumulated (once every epoch). Early in the quantization training, the rough estimate gained by true \ac{SGD} (large $r$ value) is good enough for both algorithms, as the quantum fits are anyway not optimized yet. By the end of the training, where precise fitting is needed, $r=1$, so that updates happen on fully accumulated results, basically turning the optimization into (non-stochastic, regular) Gradient Descent.
			
			To realize accumulation, for both algorithms when not updating, we maintain a target tensor $T$ (target) of shape $(k, d)$, and a corresponding weight tensor $W$ of shape $(k)$. When updating, tensor $T$ is forward propagated as input through the layers, and the resulting masked distance tensor $D_m$ is summed to create a final loss value $l$. This $l$ is then backpropagated through the distance selection and masking layers to update the quantum centroids.
			
			For \kmeans{}, $T$ holds the running average of assigned training points for each quanta since the last update, while $W$ holds the number of training points that contributed to the running average. When forward propagating, in order to find which training points are assigned to which quanta, the mask $M$ from the distance selection layer can be used. For each column (quanta) in $M$, the position where a rows contains the value $1$, the value from the corresponding position in $B$ is used to calculate a batch and quantum-wide mean $T'$ $(k, d)$. The number of points that make up each mean is can be computed by summing each column in $M$, and is stored in a temporary tensor $W'$ $(k)$. Each row in tensor $T$ is then updated according to:
			\begin{equation}
				T[i] = \frac{W[i] * T[i] + W'[i] * T'[i]}{W[i] + W'[i]},
			\end{equation}			
			where $[i]$ refers to the corresponding subset along the first tensor dimension, i.e. row or single value. $W$ is updated according to $W = W + W'$. 
			
\begin{figure}[h]
				\vspace{-0.2in}
				\centering
				\subfloat[\kmeans{}]{
					\includegraphics[width=0.46\linewidth]{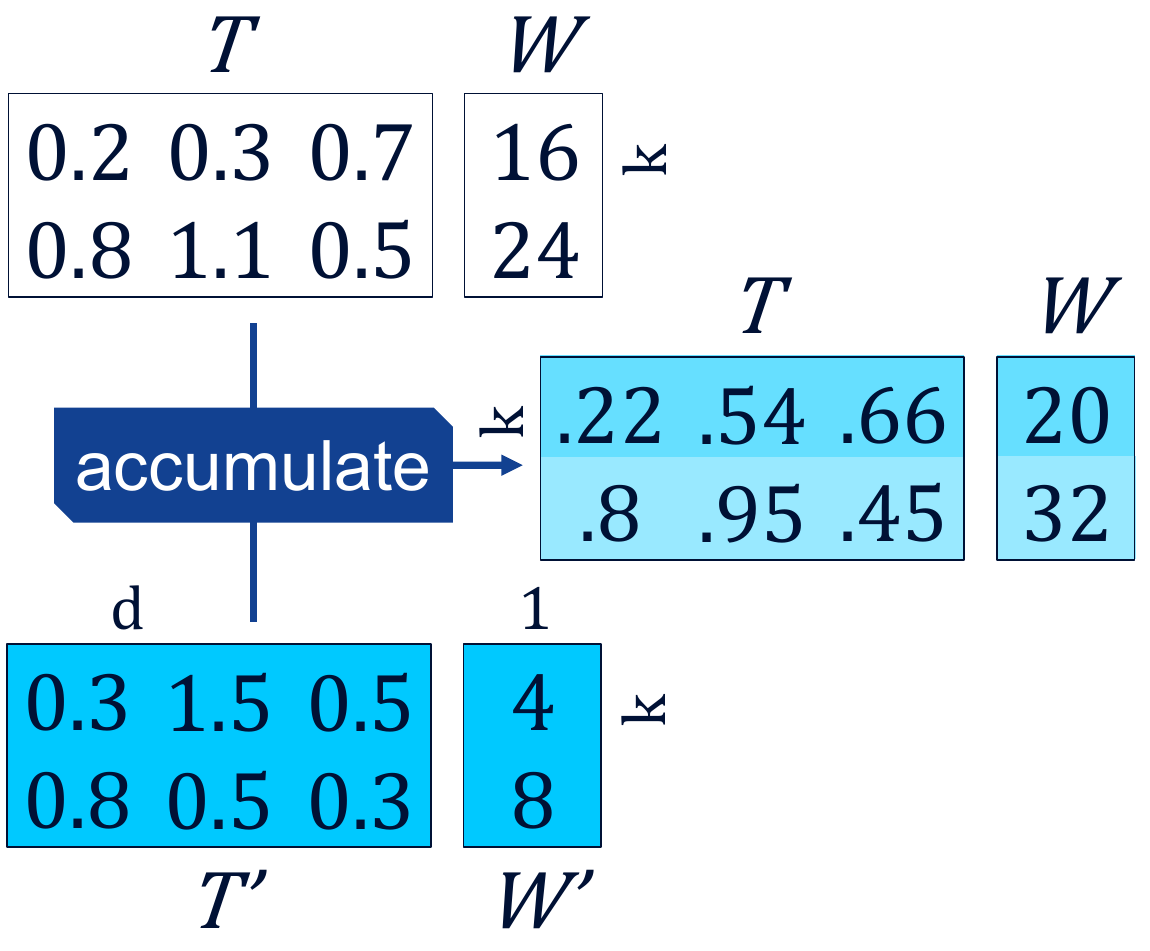}
				}
				\subfloat[\ac{BSQ}]{
					\includegraphics[width=0.46\linewidth]{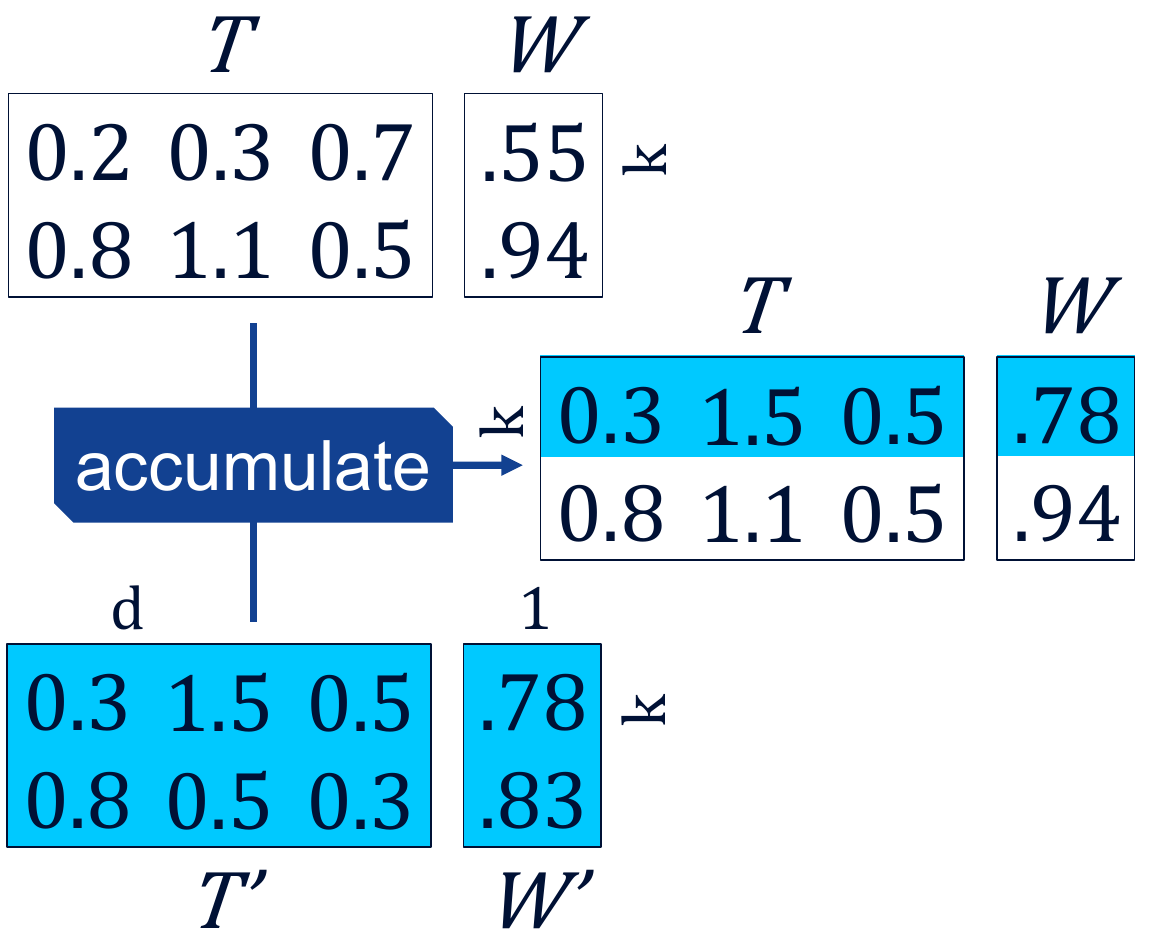}
				}
				\caption{Accumulation examples}
				\label{fig:accumulate}
				\vspace{-0.1in}
			\end{figure}

			For \ac{BSQ}, $T$ holds the so far found farthest training point for each quanta, while $W$ holds the distance of said point to the corresponding quantum, while $T'$ and $W'$ are equivalent tensors for the current batch. Both can be generated by selecting the row from $B$ where (for each column) the value in $D_m$ was the largest; the rows from $B$ make up $T'$, while the largest values from $D_m$ make up $W'$. Now, the row $T[i]$ is overwritten with $T'[i]$, if $W'[i] > W[i]$. Similarly, $W[i]$ is also overwritten with $W'[i]$ in this case. The process of accumulation for both \kmeans{} and \ac{BSQ} can be seen on Fig. \ref{fig:accumulate}. The values of tensor $W$ are set to $0$ after each update for both algorithms, to restart the accumulation of targets in $T$.
			
			Both the accumulation and the use of \ac{SGD} are critical components for the correct functioning of \ac{BSQ}. Accumulation makes it possible to find the true training-set-wide farthest points, while \ac{SGD} replaces the fitting of minimal bounding spheres present in the original \ac{BSQ}. As an illustration of how this works; when quantum centroids end up in the middle between two farthest points, \ac{SGD} moves the quantum centroid towards one of the farthest points in one iteration, and towards the other in the next, approximating the move towards the center of the minimal bounding sphere. Usually, by the end of the training, the learning rate is low, so the noise caused by this jitter is barely noticeable in the quantization.

	\section{Related Work and Evaluation}
	
		Large amount of research has been done with the aim of speeding up the original \kmeans{} algorithm. Among many ideas, the two most frequently utilized are the use of indexing schemes (such as kd-trees) to speed up search for the closest quanta \cite{kmeansKD}, and the use of the triangle inequality to avoid the calculation of distances whenever possible \cite{kmeansFaster}\cite{kmeansTriangle}. Although algorithmically faster, these ideas are complicated to realize in the massively parallel processing environment of a \ac{GPU}. A \ac{GPU} can potentially run parallel thread executions numbering in the ten thousands, for which the duplication of indexing structures (such as kd-trees) would be infeasible. The use of triangle inequality is not as simple to dismiss, and there have been successful implementations of this scheme that utilize a \ac{GPU} \cite{kmeansCuda}. However, the logic is very complex, and the speedup is heavily dependent on data ordering/structure.
				
		\ac{BSQ} is not a well-known algorithm, and as such, has not seen research regarding speedup so far. However, the core of the originally proposed \ac{BSQ}, the fitting of the minimal bounding sphere is a well-researched subject. Fischer's algorithm \cite{fischer} is the so far found quickest method, however, due to many aspects, it is hard to implement to run on a \ac{GPU}.
		
		Our \kmeans{} and \ac{BSQ} implementation was written in the Python\footnote{https://www.python.org/} language, utilizing the PyTorch library for \ac{GPU} acceleration. For reference, we chose a readily available \kmeans{} implementation that fits into this software environment from the SciPy\footnote{https://www.scipy.org/} library (\textit{scipy.cluster.vq.kmeans2}), which utilizes multi-threaded \ac{CPU} execution, but no \ac{GPU} acceleration. The evaluation was run on a system with an AMD Ryzen Threadripper 1920X 12-core \ac{CPU} with $64$ GB of memory, and an Nvidia GeForce 1080 Ti \ac{GPU} with $12$ GB of memory.
		
		Overall, training the \kmeans{} and \ac{BSQ} algorithms are not computationally heavy tasks (compared to for example training a state-of-the-art \ac{DNN}), so moving batches of data to and from the \ac{GPU} can create a significant overhead on smaller datasets. Conversely, large datasets usually do not fit into the limited memory of a \ac{GPU}, and the user is forced to keep the dataset in \ac{CPU} memory and process it batch-by-batch on the \ac{GPU}. In our evaluation, both batched and non-batched versions of the algorithms were measured, with the batched versions denoted as \ac{BSQ}$_b$ and \kmeans{}$_b$. The results can be seen on Table \ref{tab:times}, where $k$ denotes the number of quanta fitted, and $n$ the number of training points. The training points were randomly generated from a $d$-dimensional normal distribution with $0$ mean and $1$ standard deviation. Each algorithm was run for a $100$ epochs.
		
		\definecolor{one}{RGB}{193,213,247}
		\definecolor{two}{RGB}{141,177,255}
		\definecolor{three}{RGB}{217,221,227}
		\definecolor{four}{RGB}{180,187,198}
		
		\newcommand{\hlone}[1]{\cellcolor{one}\textcolor{black}{#1}}
		\newcommand{\hltwo}[1]{\cellcolor{two}\textcolor{black}{#1}}
		\newcommand{\hlthree}[1]{\cellcolor{three}\textcolor{black}{#1}}
		\newcommand{\hlfour}[1]{\cellcolor{four}\textcolor{black}{#1}}
		
		\begin{table}[t]
			\centering
			\renewcommand*{\arraystretch}{1.2}
			\setlength\tabcolsep{4pt}
			\caption{Measured runtimes [seconds]}
			\label{tab:times}
			\begin{tabular}{|l|l|l|r r r r r|l}
				\cline{1-3}
				k						& n							& d			& \ac{BSQ}$_b$	& \kmeans{}$_b$ & \ac{BSQ}	& \kmeans{}	& SciPy		&		\\
				\cline{1-8}
				\multirow{9}{*}{$32$}	& \multirow{3}{*}{$10^3$}	& $10$		& $16.29$		& $16.12$		& $0.14$	& $0.15$	& $0.02$	&		\\
				\cline{3-3}
										& 							& $10^2$	& $16.53$		& $16.58$		& $0.11$	& $0.11$	& $0.05$	&		\\
				\cline{3-3}
										&							& $10^3$	& $16.93$		& $17.18$		& $0.41$	& $0.46$	& $0.49$	&		\\
				\cline{2-8}
										& \multirow{3}{*}{$10^4$}	& $10$		& $19.76$		& $19.93$		& $0.25$	& $0.16$	& $0.15$	&		\\
				\cline{3-3}
										&							& $10^2$	& $20.11$		& $20.34$		& $0.48$	& $0.45$	& $0.56$	&		\\
				\cline{3-3}
										&							& $10^3$	& $22.79$		& $23.45$		&\hlone{$3.06$}	&\hlone{$3.81$}	&\hlone{$5.04$}	&\hlone{} 						\\
				\cline{2-8}
										& \multirow{3}{*}{$10^5$}	& $10$		& $43.85$		& $46.34$		&\hlone{$1.33$}	&\hlone{$0.64$}	&\hlone{$1.42$}	&\hlone{}						\\
				\cline{3-3}
										& 							& $10^2$	& $45.87$		& $48.79$		&\hlone{$3.74$}	&\hlone{$3.77$}	&\hlone{$5.77$}	&\hlone{\multirow{-3}{*}{$(1)$}}	\\
				\cline{3-3}
										& 							&$10^3$		&\hlthree{$69.97$}		&\hlthree{$75.22$}		& $-$		& $-$		&\hlthree{$49.54$}	&\hlthree{$(3)$}	\\
				\cline{1-8}
				\multirow{9}{*}{$512$}	& \multirow{3}{*}{$10^3$}	& $10$		& $16.96$		& $17.35$		& $0.22$	& $0.19$	& $0.18$	&		\\
				\cline{3-3}
										& 							& $10^2$	& $17.29$		& $17.39$		& $0.63$	& $0.74$	& $0.44$	&		\\
				\cline{3-3}
										& 							& $10^3$	& $21.67$		& $22.88$		&\hltwo{$5.18$}	&\hltwo{$6.44$}	&\hltwo{$3.29$}	&\hltwo{$(2)$}	\\
				\cline{2-8}
										& \multirow{3}{*}{$10^4$}	& $10$		& $18.92$		& $19.60$		& $1.01$	& $1.04$	& $2.02$	&		\\
				\cline{3-3}
										& 							& $10^2$	&\hlfour{$24.26$}		&\hlfour{$25.08$}		& $-$		& $-$		&\hlfour{$4.42$}	&\hlfour{$(4)$}		\\
				\cline{3-3}
										& 							& $10^3$	&\hlthree{$70.19$}		&\hlthree{$82.23$}		& $-$		& $-$		&\hlthree{$29.26$}	&\hlthree{}		\\
				\cline{2-8}
										& \multirow{3}{*}{$10^5$}	& $10$		&\hlthree{$42.39$}		&\hlthree{$44.93$}		& $-$		& $-$		&\hlthree{$16.87$}	&\hlthree{}		\\
				\cline{3-3}
										& 							& $10^2$	&\hlthree{$83.01$}		&\hlthree{$99.08$}		& $-$		& $-$		&\hlthree{$40.05$}	&\hlthree{}		\\
				\cline{3-3}
										& 							& $10^3$	&\hlthree{$546.09$}		&\hlthree{$669.21$}		& $-$		& $-$		&\hlthree{$296.92$}	&\hlthree{\multirow{-4}{*}{$(3)$}}	\\
				\cline{1-3}
			\end{tabular}
			\vspace{-0.2in}
		\end{table}
		
		Generally, the non-batched versions of \kmeans{} and \ac{BSQ} are quite competitive with the SciPy implementation, even winning in cases of low $k$ but high $n$ or $f$ values (highlighted as 1). The SciPy implementation probably incorporates some form of speedup scheme (such as kd-trees, but there is no reference in the documentation), as its runtime does not scale linearly with $k$, and so it wins out for large values of $k$ (2). Dashes denote data sizes where the data and the network together no longer fit into the \ac{GPU} memory, and as such, non-batched versions of our algorithms were no longer feasible to run.
		
		Batched versions used a batch size of $512$. With this batch size, small datasets incurred such a heavy overhead that \ac{BSQ}$_b$ and \kmeans{}$_b$ could run several magnitudes slower than the SciPy or their non-batched counterparts. However, at the point where using the non-batched versions becomes infeasible, the batched versions are only $2x-3x$ slower than the SciPy implementation (3), except in the single worst case of (4). These runtimes still contain a considerable amount of overhead from data transfers, which means that if the algorithm is a part of a larger processing pipeline, these overheads could be hidden to result in an even shorter partial runtime.

	\section{Conclusion}
	
		As a follow-up to our previous paper, we have shown an implementation of the \kmeans{} and \ac{BSQ} algorithms, using common tensor operations that can be found in any state-of-the-art \ac{DL} framework. We have evaluated our implementation, which utilizes \ac{GPU} acceleration, against a commonly used, \ac{CPU}-based implementation. We have found that our simple approach measures up against the more complex algorithm, sometimes even producing shorter runtimes.
		
		Our proposed implementation is easy to realize in many of the currently popular \ac{DL} frameworks, lending itself to be used as a processing step in a larger machine learning pipeline. Additional benefits from using a familiar framework, such as simplicity, ease-of-use, or modifiability are not measurable, but are nevertheless invaluable qualities to us researchers, and should also be kept in mind. All-in-all, we think that the implementations shown here, especially for \ac{BSQ}, is a worthy addition to the toolset of any network automation researcher.
		
	\clearpage

	\bibliographystyle{IEEEtran}
	\bibliography{bibliography}{}
	
	
	
\end{document}